\documentclass{article}

\usepackage{arxiv}

\usepackage{algorithm}        
\usepackage{algorithmicx}     
\usepackage{algpseudocode}    
\usepackage{amsfonts}         
\usepackage{amsmath,amssymb,amsfonts} 
\usepackage{amsthm}           
\usepackage{appendix}         
\usepackage{array}            
\usepackage{booktabs}         
\usepackage{datatool}         
\usepackage{doi}              
\usepackage{fontenc}          
\usepackage{graphicx}         
\usepackage{hhline}           
\usepackage{hyperref}         
\usepackage{inputenc}         
\usepackage{lipsum}           
\usepackage{listings}         
\usepackage{longtable}        
\usepackage{manyfoot}         
\usepackage{mathrsfs}         
\usepackage{microtype}        
\usepackage{multirow}         
\usepackage{multicol}         
\usepackage{natbib}           
\usepackage{nicefrac}         
\usepackage{paralist}         
\usepackage{subcaption}       
\usepackage{svg}              
\usepackage{textcomp}         
\usepackage{tikz}             
\usepackage{url}              
\usepackage{xcolor}           
\usepackage{cleveref}         

\title{Deep Learning Based Crime Prediction Models: \\Experiments and Analysis}

\usepackage{authblk}

\author[1]{%
	\hspace{1mm}Rittik Basak Utsha%
}
\author[1]{%
	\hspace{1mm}Muhtasim Noor Alif%
}
\author[2]{%
	\hspace{1mm}Yeasir Rayhan%
}
\author[1]{%
	\hspace{1mm}Tanzima Hashem%
}
\author[1]{%
	Mohammad Eunus Ali%
}
\affil[1]{Department of Computer Science and Engineering, Bangladesh University of Engineering and Technology, ECE Building, Dhaka, 1000, Bangladesh}
\affil[2]{
Purdue University, West Lafayette, IN
}

\renewcommand{\shorttitle}{DL crime prediction models}

\hypersetup{
pdftitle={Deep Learning Based Crime Prediction Models: Experiments and Analysis},
pdfauthor={Rittik Basak Utsha, Muhtasim Noor Alif, Yeasir Rayhan, Tanzima Hashem, Mohammad Eunus Ali},
pdfkeywords={Crime prediction, Deep learning, Experimental analysis, Performance evaluation},
}

\begin{document}
\maketitle

\begin{abstract}
	Crime prediction is a widely studied research problem due to its importance in ensuring safety of city dwellers. Starting from statistical and classical machine learning based crime prediction methods, in recent years researchers have focused on exploiting deep learning based models for crime prediction. Deep learning based crime prediction models use complex architectures to capture the latent features in the crime data, and outperform the statistical and classical machine learning based crime prediction methods. However, there is a significant research gap in existing research on the applicability of different models in different real-life scenarios as no longitudinal study exists comparing all these approaches in a unified setting. In this paper, we conduct a comprehensive experimental evaluation of all major state-of-the-art deep learning based crime prediction models. Our evaluation provides several key insights on the pros and cons of these models, which enables us to select the most suitable models for different application scenarios. Based on the findings, we further recommend certain design practices that should be taken into account while building future deep learning based crime prediction models.
\end{abstract}


\section{Introduction}
\label{intro}
\begin{figure*}
    \centering
    \includegraphics[width=\textwidth]{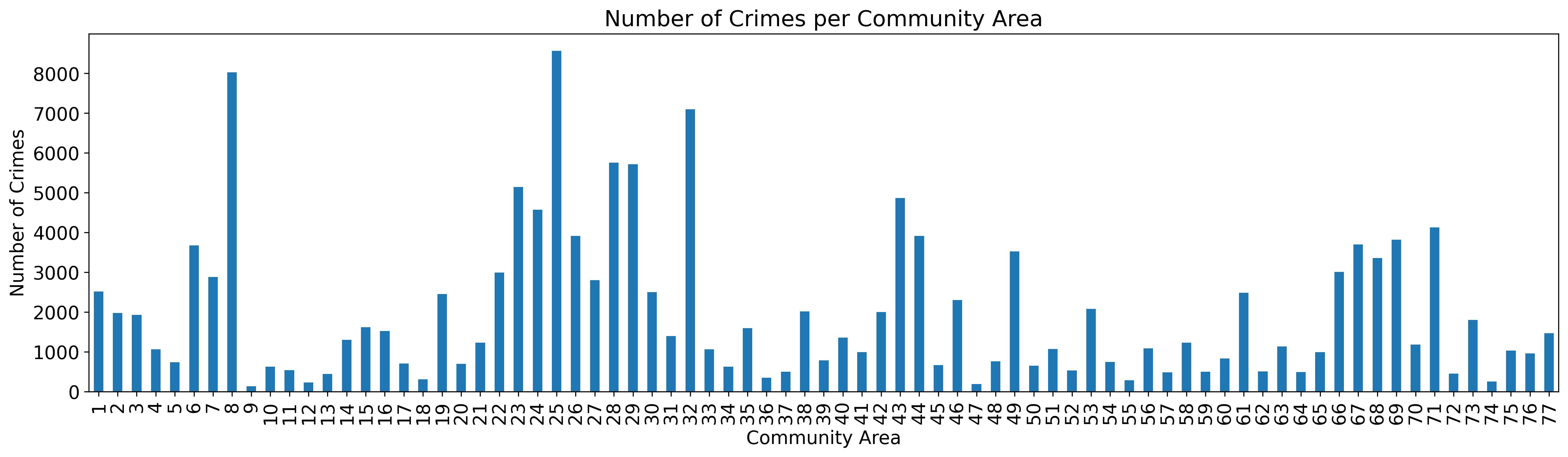} \\[1em] 
    \includegraphics[width=\textwidth]{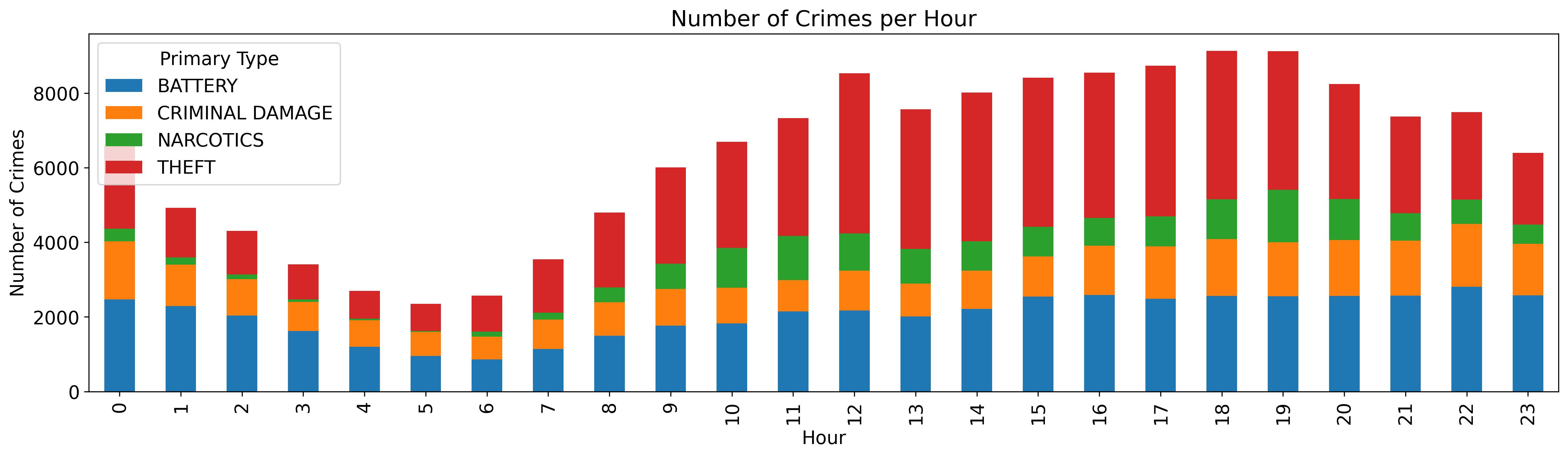} \\[1em]
    \includegraphics[width=\textwidth]{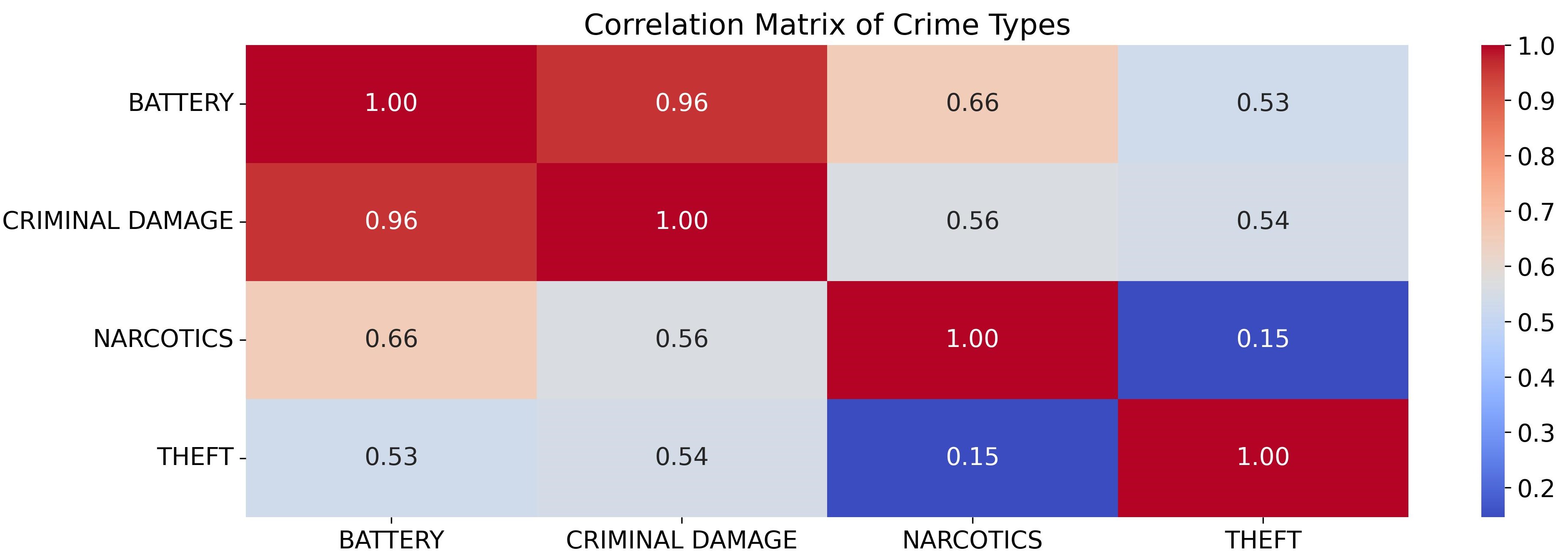}
    \caption{(a) Spatial Dependency: Crimes can exhibit different patterns for different regions. The communities of Chicago exhibit a wide range of number of crimes in 2019. (b) Temporal Dependency: Crime frequency can be different on different time of the day. The earlier hours of a day have less occurrences of crimes than later in the day in our dataset. (c) Categorical Dependency: One crime can be dependent on another. A heatmap depicting the correlations between the crime categories in our dataset is shown.}
    \label{fig:dependencies}
\end{figure*}

Crime prediction problem has been extensively studied in the literature. Early prediction of crime helps the security enforcing authorities to take preventive measures. 
The task of crime prediction involves analyzing a city, region, or space to forecast the likelihood of a specific type of crime occurring or the number of crimes that might happen, based on past crime data in that area. Modeling crime patterns in a target area is a complex task because various factors can influence them. For instance, locations like shopping malls, businesses, and universities may experience higher crime rates due to the large number of people frequenting these areas. This is referred as the spatial dependency of crimes. In our paper, we refer to these places as points of interest (POIs). Additionally, crime rates in a region can vary depending on the time of day, month, or year. Urban areas with poor lighting and security measures at night may attract criminals. This dependency of crime on time is referred as temporal dependency. Moreover, crime patterns in one region can be affected by neighboring regions; one type of crime may increase the likelihood of another. For instance, a robbery might lead to a hit-and-run incident. The correlations of one crime to another is referred as categorical dependency. These dependencies are illustrated in Figure \ref{fig:dependencies}. These factors significantly impact crime rates in a region, making it essential for state-of-the-art models to consider them when modeling crime patterns.
Researchers have developed statistical and classical machine learning based crime prediction methods~\cite{arima, dtr, gbdt, stresnet, att-rnn} in the last decade. To further improve the prediction accuracy, recent research have considered developing deep learning based crime prediction models: DeepCrime (DC)~\cite{deepcrime}, MIST~\cite{mist}, CrimeForecaster (CF)~\cite{crimeforecaster}, HAGEN~\cite{hagen}, ST-SHN~\cite{stshn} and ST-HSL~\cite{sthsl}, AIST~\cite{aist} for crime prediction. As these approaches use a wide variety of settings in their experiments, and no longitudinal study exists in an unified environment, it is quite challenging to identify the most suitable model for a particular scenario, which limits the applicability of these models in a real-life environment. 
To overcome this limitation of existing research, we perform an experimental analysis of the proposed seven deep learning based models, and identify the key insights and the pros and cons of these state-of-the models, which enable us to find the most suitable model in different real-life settings.

The motivation behind our experimental study are as follows.

\textbf{Missing competitors (M1).} The deep learning based crime prediction models under consideration compare themselves with one or two other deep learning based models in experiments (Table~\ref{table:comp_table}). For example, AIST was compared with MIST and DeepCrime, HAGEN  was compared with MIST and CrimeForecaster, and ST-SHN  was compared with DeepCrime. However, all of the models claim their superiority over statistical and classical machine learning based crime prediction methods. As a result, there is a need to compare all of the deep learning based models to know their actual performance. 

\textbf{Missing experiments (M2).} The deep learning based crime prediction models only show their performance for different crime categories. They fail to answer questions like how does a model perform if crime data density or the region area or the time interval for which the crime occurrence is predicted vary. Some deep learning based crime prediction models~\cite{sthsl, stshn, aist} consider predicting the number of crime occurrences, whereas others~\cite{deepcrime, mist, crimeforecaster, hagen} predict whether a crime would occur or not for a region at a particular time interval.

\textbf{Lack of uniform experiment settings (M3).} The deep learning based crime prediction models in our study do not follow any uniform experiment settings. For example, AIST considers 4 hour time interval for crime prediction, whereas ST-SHN uses 24 hour time interval. Furthermore, the models also use different datasets in experiments.   

Though there are a few studies (~\cite{10.1007/978-3-030-14680-1_40, 10.1109/ACCESS.2020.3028420, 10.1007/s12652-023-04530-y, arXiv:2303.16310}) that compare crime prediction methods in the literature, none of them perform experimental analysis for the recent deep learning models. A study in~\cite{10.1007/978-3-030-14680-1_40} presents a comparison among Bayesian networks, random trees, and neural networks for analyzing crime patterns. In another study~\cite{10.1109/ACCESS.2020.3028420}, various classic machine learning based crime prediction algorithms, including K-nearest neighbors (KNN)~\cite{1053964}, Support Vector Machines (SVM)~\cite{vapnik1995training}, Long Short-Term Memory (LSTM)~\cite{hochreiter1997long}, and Convolutional Neural Networks (CNN)~\cite{lecun1998gradient} were compared. Both~\cite{10.1007/s12652-023-04530-y} and~\cite{arXiv:2303.16310} present surveys of existing crime prediction methods and find their trends. They do not perform any experimental analysis.


\begin{table*}[htbp]
    \centering
    \setlength{\abovecaptionskip}{-0.2pt}
    \caption{Considered baselines for deep learning based crime prediction models in experiments. (A blank cell denotes that the two models did not compare with each other, and a hyphen marked cell represents that there were no scope for comparisons as the model appeared earlier.)}
    \label{table:comp_table}
    \small
    \resizebox{\textwidth}{!}{
    
    \begin{tabular}{c|c|c|c|c|c|c|c}
        \hline
         & DC & MiST & CF & HAGEN & ST-SHN & ST-HSL & AIST \\ \hline
        DC (Huang \textit{et al.} 2018)~\cite{deepcrime} & - & - & - & - & - & - & - \\ \hline
        MiST (Huang \textit{et al.} 2019)~\cite{mist} &  & - & - & - & - & - & - \\ \hline
        CF (Sun \textit{et al.} 2019)~\cite{crimeforecaster} &  & \checkmark & - & - & - & - & - \\ \hline
        HAGEN (Wang \textit{et al.} 2021)~\cite{hagen} &  & \checkmark & \checkmark & - & - & - & - \\ \hline
        ST-SHN (Xia \textit{et al.} 2021)~\cite{stshn} & \checkmark &  &  &  & - & - & - \\ \hline
        ST-HSL (Li \textit{et al.} 2022)~\cite{sthsl} & \checkmark &  &  &  & \checkmark & - & - \\ \hline
        AIST (Rayhan \& Hashem 2023)~\cite{aist} & \checkmark & \checkmark &  &  &  &  & - \\ \hline
    \end{tabular}
    }
\end{table*}

In this paper, we perform a comprehensive experiment analysis of deep learning based crime prediction models under a unified experiment setting. Specifically, we evaluate the performance of the models by varying the size, crime data density of the target region. We further evaluate the models while varying the granularity of the temporal precision of the predictions. With these categorization, we aim to do an exhaustive search over all possible scenarios in terms of the geographical property, sparsity of crime data, temporal granularity and evaluate the models under different scenarios. This approach further helps us gain insights into the most suitable neural network architecture for different scenarios.

Our \textbf{\textit{key findings}} include when crime data is very sparse, models (e.g., AIST) that attempt to capture the interaction between external features and crime data tend to perform better for regression task. However, as the sparsity of crime data tends to decrease, models (e.g., CrimeForecaster, HAGEN) benefit from utilizing the information from regions with similar crime profile. We observe the same phenomenon when we vary the area of the target regions due to the positive correlation between the area of a target region and its crime occurrences. Contrary to regression for classification task, we find that models that explicitly capture the different temporal trends of crime data, i.e., recent, daily and weekly tend to perform better for classification task across all possible scenarios presented in our experimental setting.

Through a detailed experimental study and critical comparative analysis of the results, we \textbf{\textit{pinpoint eight critical questions}} that are of utmost importance for any crime prediction system and answer them one by one. Refer to Section~\ref{sec:findings} for the questionnaire list. These include figuring out (i) the best performing model across all scenarios for regression (\textbf{Q1}) and classification task (\textbf{Q2}), (ii) the best performing model under specific scenarios, i.e., precise temporal precision of prediction (\textbf{Q3}), very sparse crime data (\textbf{Q4}), (iii) evaluating the necessity (\textbf{Q5}) and utilization techniques (\textbf{Q6}) of external features, (iv) impact of capturing spatial dependencies (\textbf{Q7}), and (v) the desired model characteristics of a crime prediction model for regression and classification task (\textbf{Q8}).

Based on these findings, we further \textbf{\textit{recommend}} certain design practices that should be taken into account while building deep learning based crime prediction models in terms of (i) the modeling of spatio-temporal correlation (\textbf{R1}) of crime data, (ii) prediction in the absence of external features (\textbf{R2}), (iii) utilization schemes of external features when present (\textbf{R3}), (iv) desiderata of a crime prediction model for regression task (\textbf{R4}), (v) desiderata of a crime prediction model for classification task (\textbf{R5}). 



In summary we have made the following contributions.
\begin{compactitem}
\item We have done a critical analysis of different deep learning architectures and components used in these crime prediction approaches, where we have specifically identified commonness and differences of these architectures. We have also analyzed the models in terms of different characteristics such as used crime data features, considered spatial and non-spatial neighborhood, and prediction types (Section~\ref{sec:lit_review}).


 \item We have performed a comprehensive experiment analysis of deep learning based crime prediction models in a unified experimental setting. We have evaluated the performance of the models by varying the crime data density, the area of the region and the time interval for which the crime occurrence is predicted to find the best model for each scenario (Section~\ref{sec:experiments})

 \item We have summarized the findings with respect to answers of key questionnaire that can be vital in choosing crime prediction models (Section~\ref{sec:findings}) and provide a set of recommendations (Section~\ref{sec:reco}).
\end{compactitem}


\section{Deep Learning Models for Crime Prediction}
\label{sec:lit_review}

Recently deep learning models have become popular in many domains like image processing, speech recognition and natural language processing. Deep learning models have also been recently used to capture the non-linear spatio-temporal dependencies of crime data for better crime prediction performance~\cite{deepcrime,hagen, aist}. In this section, we first present an overview of the architecture of seven major deep learning based crime prediction models: DeepCrime~\cite{deepcrime}, MIST~\cite{mist}, CrimeForecaster~\cite{crimeforecaster}, HAGEN~\cite{hagen}, ST-SHN~\cite{stshn} and ST-HSL~\cite{sthsl}, AIST~\cite{aist}, that we have selected for our experimental evaluation (Section~\ref{ssec:modelarch}).
We then critically analyze those models to find similarities and dissimilarities (Section~\ref{ssection:comp}). 

\begin{figure*} 
\centering 
\includegraphics[width=1\textwidth]{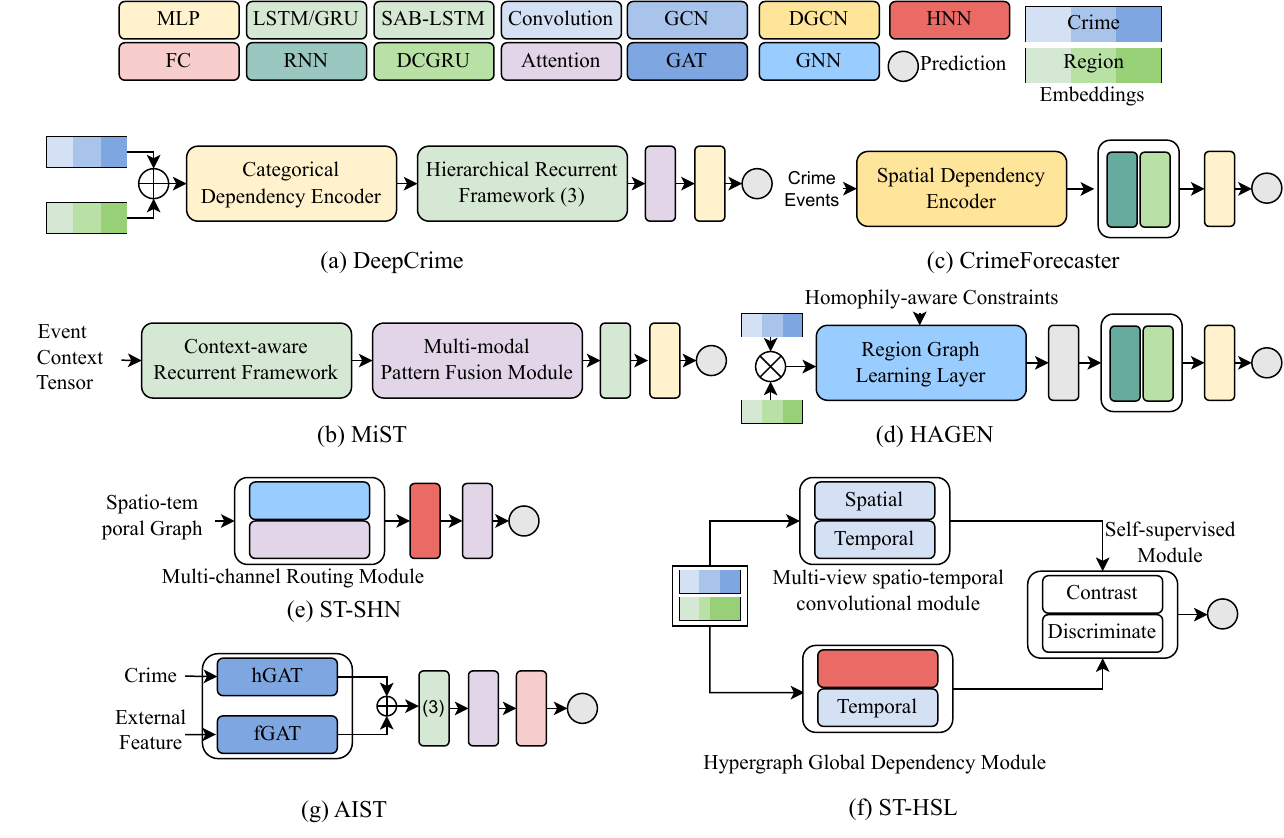}
\caption{The Architectural Synopses of Deep Learning Based Crime Prediction Models. (MLP: Multi-Layer Perceptron, FC: Fully Connected NN, LSTM: Long Short-Term Memory, GRU: Gated Recurrent Unit, RNN: Recurrent Neural Network, SAB-LSTM: Sparse Attention-Based LSTM, DCGRU: Diffusion Convolution GRU, GCN: Graph Convolutional Network, GAT: Graph Attention Network, DGCN: Diffusion GCN, GNN: Graph Neural Network, HNN: Hypergraph Neural Network)}
\label{fig:archs}
\end{figure*}

\subsection{Model Architectures}
\label{ssec:modelarch}
In this section, we discuss deep learning architectures used in different crime prediction models. The high-level diagram  of these architectures is shown in Figure~\ref{fig:archs}.
\subsubsection{DeepCrime}
\label{ssec:deepcrime}
DeepCrime~\cite{deepcrime} is one of the first models to employ deep learning for crime prediction. The model claims to capture the dynamic patterns of crime and their correlation with other influencing data (POI, anomaly) across different time steps.

Figure~\ref{fig:archs}a depicts the high level architecture of DeepCrime. 
In its architecture, the intra-region and intra-crime correlations are first embedded, and these embeddings are fed through a MLP to encode the region-category interactions in a weighted vector. The POI information is used while creating the region embedding. Then a hierarchical recurrent framework with three Gated Recurrent Units (GRU) \cite{gru} is used to encode the temporal dependencies of crime sequence, anomaly sequence, and their inter-dependencies. Next an attention layer \cite{attention} is introduced that models the influence of past crimes for the prediction of future crimes. Finally, a MLP network maps the learned vectors to output the crime probability. DeepCrime authors compare their model with traditional ML models such as SVR~\cite{svr}, ARIMA~\cite{arima}, LR~\cite{lr}, and GRU~\cite{gru} with NYC crime data~\cite{nyc_dataset} of 2014.



\subsubsection{MiST: Multi-View Deep Spatial-Temporal Network}
\label{ssec:mist}


MiST~\cite{mist} is a model designed to predict the occurrence of spatial-temporal abnormal events, such as crimes, urban anomalies, etc.
Figure \ref{fig:archs}b shows the overview of the MiST architecture. MiST first divides the target city with a grid-based map segmentation. With the data mapped into the cells of this grid, MiST first employs a Long Short-Term Memory (LSTM)~\cite{lstm} based network, named "Context Aware Recurrent Framework", to encode the time-dependent nature of the data. Then the inter-region and cross-category correlations are captured through the use of an attention mechanism, named "Multi-Modal Pattern Fusion Module." After that, the complex interactions between the spatial-categorical fusion module and temporal recurrent module are integrated through another recurrent module, named "Conclusive Recurrent Network". Finally, the output of the recurrent module are fed through a MLP network to generate the occurrence probabilities. 


The authors experimented their model with NYC Crime data~\cite{nyc_dataset}, NYC Urban Anomaly Data and Chicago Crime Data~\cite{dataset}. 

MiST is compared against various traditional ML models such as SVR~\cite{svr}, ARIMA~\cite{arima} and some early deep learning models such as ST-RNN~\cite{st_rnn}, RDN~\cite{RDN}, and ARM~\cite{arm}.



\subsubsection{CrimeForecaster}
\label{ssec:crimeforecaster}

Crimeforecaster~\cite{crimeforecaster} authors argue that the spatial nature of crime depends on the temporal nature, i.e. a community can have different crime patterns depending on the time of the year, month, week or day. They claim that traditional methods process the spatial and temporal information separately, hence neglect the spatio-temporal dependency. Crimeforecaster models this spatio-temporal dependency using a Diffusion Convolution module. 
Figure~\ref{fig:archs}c shows a high level diagram of CrimeForecaster. CrimeForecaster represents the neighborhoods and learns their correlations through a graph, and then uses diffusion convolution on that graph to predict crimes. CrimeForecaster uses a DCGRU (Diffusion Convolutional Gated Recurrent Units)~\cite{dcrnn} encoder to learn the complex intra and inter-region correlations across the previous time slots. The diffusion convolution operation models the spreading nature of crimes to nearby and similar regions with time.





\subsubsection{HAGEN: Homophily-Aware Graph Convolutional Recurrent Network}
\label{ssec:hagen}



HAGEN \cite{hagen} authors claim that two areas may exhibit similar crime patterns if they share some common traits such as proximity in geographical distance or similar points of interest. Instead of creating individualized graphs for every instance, HAGEN suggests identifying similar regions in the graph using adaptive graph learning.

Figure \ref{fig:archs}d shows the high level diagram of HAGEN. It first applies a region-crime dependency encoder, which under the hood learns the graph adaptively, using a homophily-aware constraint. Crime and region embeddings are utilized to jointly capture the interactions between regions and crimes. Region embeddings are created based on geographical distances and points of interests. After this layer, the temporal dependency of the crimes are caputred using a diffusion convolutional recurrent module, namely DCGRU, just like Crimeforecaster. The final predictions is obtained by an MLP-based decoder.


The authors used almost similar experimental setup as CrimeForecaster~\cite{crimeforecaster}. They claim that HAGEN outperforms CrimeForecaster and other competitive models. 

\begin{table*}[]
\centering
\caption{Components used by the models to capture various aspects of crime data}
\label{tab:data_feature_vs_arch}
\begin{tabular}{c|ccc}
\hline
\textbf{Model} &
  \multicolumn{1}{c|}{\textbf{\begin{tabular}[c]{@{}c@{}}Categorical\\ Dependency\end{tabular}}} &
  \multicolumn{1}{c|}{\textbf{\begin{tabular}[c]{@{}c@{}}Spatial\\ Dependency\end{tabular}}} &
  \multicolumn{1}{c|}{\textbf{\begin{tabular}[c]{@{}c@{}}Temporal\\ Dependency\end{tabular}}}  \\ \hline
DC &
  \multicolumn{2}{c|}{MLP} &
  \multicolumn{1}{c}{GRU and Attention} \\ \hline
MiST &
  \multicolumn{2}{c|}{Attention} &
  \multicolumn{1}{c}{LSTM}
   \\ \hline
CF &
  \multicolumn{3}{c}{DCGRU} 
   \\ \hline
HAGEN &
  \multicolumn{2}{c|}{MLP, Graph Learning} &
  \multicolumn{1}{c}{DCGRU} 
  \\ \hline
ST-SHN &
  \multicolumn{1}{c|}{\begin{tabular}[c]{@{}c@{}}Message Passing,\\ Attention\end{tabular}} &
  \multicolumn{1}{c|}{Hypergraph} &
  \multicolumn{1}{c}{\begin{tabular}[c]{@{}c@{}}Message Passing,\\ Attention\end{tabular}} 
   \\ \hline
\multirow{2}{*}{ST-HSL} &
  \multicolumn{3}{c}{Convolution (local)} 
   \\ \cline{2-4} 
 &
  \multicolumn{3}{c}{Hypergraph (global)} 
   \\ \hline
AIST &
  \multicolumn{1}{c|}{hGAT} &
  \multicolumn{1}{c|}{hGAT, fGAT} &
  \multicolumn{1}{c}{SAB-LSTM} 
   \\ \hline
\end{tabular}
\end{table*}

\subsubsection{ST-SHN: Spatial-Temporal Sequential Hypergraph Network for Crime Prediction with Dynamic Multiplex Relation Learning}
\label{ssec:sthsn}

Xia et al.~\cite{stshn} suggest using a graph-based method for message passing among different regions for different crime categories, incorporating hypergraph learning to address spatial and temporal changes within a broad context.

ST-SHN first generates a region graph representing the regions and their geographical adjacencies. This graph is utilized in the "Spatial Dependency Encoder" module, which captures the complex spatial dependencies among regions regarding various crime types. It incorporates  a message propagation approach called the "Multi-Channel Routing Mechanism" to model how different types of crimes influence each region. It employs a multiplex mutual attention network under the hood. To further improve cross-region learning, a hypergraph neural network named "Cross-Region Hypergraph Relation Learning" is introduced to understand the broader relationship between crimes. Temporal dependencies of crimes are handled similarly to spatial dependencies, with a network resembling the "Spatial Dependency Encoder" that spreads temporal messages within and across regions and different crime categories. The overall architecture of ST-SHN is shown in Figure~\ref{fig:archs}e.

ST-SHN has been evaluated and compared to other baseline models like ARIMA~\cite{arima}, DeepCrime~\cite{deepcrime}, DCRNN~\cite{dcrnn}, GMAN~\cite{gman} using both the NYC and Chicago crime data, and shown to have superior performance over these models.


\subsubsection{ST-HSL: Spatial-Temporal Hypergraph Self-Supervised
Learning for Crime Prediction}
\label{ssec:sthsl}

Li et al.~\cite{sthsl} claim that the problem in the state-of-the-art crime prediction models are that they all follow supervised learning methods, which require labeled data. But in real life scenario, the labels are very scarce compared to the vastness of a city. This scarcity of labeled data creates a challenge in effectively training these models effectively for real-world datasets. This is why the authors of ST-HSL proposes a dual-stage self-supervised learning paradigm.

ST-HSL encodes the spatial and temporal interactions of crimes between geographically neighboring regions through a convolutional module, named "Multi-View Spatial-Temporal Convolution", creating a local crime embedding. At the same time, it devises a hypergraph structure and employs a hypergraph-guided message passing learning framework to create a global crime embedding. A temporal convolution network is fused with the global module to inject temporal context to it. With the global and local embeddings, a "Dual-Stage Self-Supervision Learning Paradigm" is designed to tackle the challenge of data sparsity by self-supervised learning. One stage of this module generates a corrupt crime embedding and tries to discriminate between the original and corrupt graph. The other stage of the module enhances the training process by finding the contrast between the local and global embedding of crimes. The architecture of ST-HSL is illustrated in Figure~\ref{fig:archs}f.

Extensive experiments show that ST-HSL performs better than other state-of-the-art models such as ARIMA~\cite{arima}, DCRNN~\cite{dcrnn}, GMAN~\cite{gman} and deep learning models like ST-SHN~\cite{stshn}, DeepCrime~\cite{deepcrime} for crime prediction.


\subsubsection{AIST}
\label{ssec:aist}
Rayhan and Hashem~\cite{aist} claim that the current deep learning models from crime predictions are not interpretable; they also fail to address the long term temporal correlation, and do not effectively incorporate external features. Hence they propose
 an attention based deep learning model, namely AIST that uses external features like taxi flow and point of interest along with past crime data to predict crime occurrence. The high level diagram of AIST is shown in Figure~\ref{fig:archs}g.

AIST uses neighbourhood graph and hierarchical structure of the regions to capture the spatial dependency of the crime data. A variant of Graph Attention Network~\cite{gat}, hGAT to learn region crime embedding by incorporating the hierarchical information of diferent regions. Another variant, fGAT is used to embed the external features into the model. This crime embedding and feature embedding are concatenated to get the spatial embedding. The spatial representation generated at different time-steps are then fed into three Sparse Attention Based LSTM (SAB-LSTM) to capture the recent, daily and weekly trends of the crime data, thereby capturing the temporal dependency into a final weight vector that is then used to predict the crime occurrence at the next time-step.

The authors compared AIST with various other models like DeepCrime~\cite{deepcrime}, MiST~\cite{mist}, STGCN~\cite{stgcn} and found its superior performance them in terms of prediction accuracy. Also, attention weights associated with different parts of the model can be exploited to interpret its prediction.

\subsection{Comparative Analysis}
\label{ssection:comp}
In this section, we present a comprehensive comparative analysis of our crime prediction models on several key aspects: different architectural components, variants of data types employed by each model, different data features incorporated into their architectures, and the types of predictions made by these models.

\subsubsection{Sub-components of Model Architectures}
The crime prediction problem spans three dimensions: space, time, and category. Therefore, deep learning models focus on effectively modeling the interactions among these three aspects to produce accurate predictions. Table~\ref{tab:data_feature_vs_arch} summarizes the architectural synopsis of the above seven models in the context of categorical, spatial and temporal dependencies.
Previous models like DeepCrime and MiST attempt to capture spatial and categorical dependencies using MLP and attention-based layers, respectively. However, these models do not account for the graph-like relationships between regions, potentially missing important geographical semantics. CrimeForecaster, HAGEN, and AIST address this by using graph neural networks to model spatial interactions. ST-SHN and ST-HSL further utilize hypergraph neural networks to encode spatial correlations while considering the global dependency of crime among different regions. Temporal dependencies are typically captured by recurrent network variants such as GRU and LSTM, with attention mechanisms commonly used to identify important focal points in space or time.
Figure~\ref{fig:archs} depicts the block diagrams of different categories of components using different color codes.



\subsubsection{Utilization of Crime and External Datasets}
Crime prediction models typically focus on a city, using data on crimes committed within that city. The city may be divided into various regions or communities, with the number of crimes committed in each region during specific time intervals serving as the primary data for training the models. The division of the target city can vary; for instance, MiST divides the city into a grid, while CrimeForecaster considers the city as a graph with communities represented as nodes. Crimes are usually categorized into multiple types. For example, Chicago is divided into 77 different communities, and the count of various crime categories (e.g., murder, burglary, robbery, hijacking) in these communities at 4-hour intervals can form a crime dataset.  

Also, crimes in a region can be influenced by external events. Solely depending on the crime data, the models cannot capture these influences properly. So, existing deep learning models incorporate external datasets in addition to crime data of the region to enhance their prediction accuracy. These datasets provide important contextual information that help the models capture the complex factors that influence the crime pattern. The models we are using in this experimental study also utilize various external datasets to predict crime. Table~\ref{table:data_table} summarizes the datasets used by the models we are considering for this study:

\begin{table}[h]
    \centering
    \setlength{\abovecaptionskip}{-0.5pt}
    \caption{Data used by various models}
    \label{table:data_table}    
    \small
    \begin{tabular}{c|c}
        \hline
        \textbf{Model} & \textbf{Data Used} \\ \hline
        DC & Crime data, POI data, Urban anomaly data \\ \hline
        MiST & Crime data \\ \hline
        CF & Crime data \\ \hline
        HAGEN & Crime data, POI data \\ \hline
        ST-SHN & Crime data \\ \hline
        ST-HSL & Crime data \\ \hline
        AIST & Crime data, Taxi trip data, POI data \\ \hline
    \end{tabular}
\end{table}

\subsubsection{Different Types of Prediction}
Different crime prediction models adopt distinct approaches to tackle the problem. Based on the type of prediction they make, these models can be classified into two groups. The first group, DeepCrime, MiST, CF, and HAGEN, focus on predicting solely the occurrence of a crime, which is categorized as classification. 
The second group of models, ST-SHN, ST-HSL, and AIST, aim to predict the quantity of crimes that may occur during a specific time step, which is referred to as regression. Only ST-SHN consider both types of predictions. 


\section{Experimental Settings}
\label{sec:experiments}

This section provides an overview of the experimental setup for our experiments. We outline the datasets employed, evaluation criteria and parameter settings of the models for our experiments in the following subsections.

\subsection{Dataset}
\label{ssec:dataset} 
We conduct all the experiments on 2019 Chicago crime data for the following four crime categories: theft, criminal damage, battery and narcotics. In addition to the crime data, we include external feature datasets, e.g., POI information, taxi flow and urban anomaly data as required by the respective models, the details of which are presented in Table~\ref{tab:dataset_des1}. We divide the dataset such that first 8 months (January to August) are used for training, next 1 month (September) is used for validating and the data for last 3 months (October to December) are used for testing. 

We choose Chicago as our target city as most models have used Chicago crime dataset themselves and due to the availability of all the external features required to train the respective prediction models. On top of that, we claim that our categorization of Chicago communities into different groups based on the crime density can capture the properties exhibited by other target cities, i.e., Los Angeles, New York City.

\vspace{-2mm}
\begin{table}[]
\centering
\caption{Details of the Chicago datasets}
\label{tab:dataset_des1}
\centering
\begin{tabular}{c|cc}
\hline
\textbf{Datasets}                                         & \textbf{Category}     & \textbf{\#Records} \\ \hline
\multirow{4}{*}{Crime Dataset (2019)~\cite{dataset}}                     & Theft                 & 62484              \\
                                                          & Criminal Damage       & 26681              \\
                                                          & Battery               & 49513              \\
                                                          & Narcotics             & 15069              \\ \hline
\multirow{4}{*}{311 Public Service Complaint Data (2019)~\cite{311_data}} & Blocked Driveway      & 51939              \\
                                                          & Noise                 & 43542              \\
                                                          & Building/Use          & 32235              \\
                                                          & Safety                & 1415               \\ \hline
\multirow{10}{*}{POI data (2019)~\cite{POI_data}}                         & Food                  & 12532              \\
                                                          & Residence             & 4236               \\
                                                          & Travel                & 13863              \\
                                                          & Arts \& entertainment & 4780               \\
                                                          & Outdoors              & 5022               \\
                                                          & Recreation            & 3782               \\
                                                          & Education             & 5646               \\
                                                          & Nightlife             & 21014              \\
                                                          & Professional          & 18402              \\
                                                          & Shops and event       & 47                 \\ \hline
\multirow{2}{*}{Taxi Trip Data (2019)~\cite{taxi}}                    & Taxi inflow           & 14557888           \\
                                                          & Taxi outflow          & 14552209           \\ \hline
\end{tabular}
\end{table}

\vspace{-2mm}

\subsection{Parameter Settings}
Refer to Table~\ref{tab:param} for the parameter settings of the seven models under different scenarios.
\begin{table*}[thbp]
\centering
\setlength{\abovecaptionskip}{-0.5pt}
\caption{Parameter settings for the models(B=Batch size, A=Attention Dimension size, L=\#MLP layers, E=Embedding size, lr=Learning rate, Hsd=Dimension of hidden state, Ep=\#Epochs, Rl=\#RNN layers, Ds=Diffusion step, Drl=Dimension of RNN layers, Dr=Decay rate, Ss=Subgraph size, Sr=Saturation rate, H=\#Hyperedges, Ks=Kernel size, Cl=\#CNN layers, Rt=\#Recent Timesteps, Hsl=Hidden state in SAB-LSTM)}

\label{tab:param}
\begin{tabular}{c|c|cc}
\hline
\multirow{2}{*}{\textbf{Model}} & \multirow{2}{*}{\textbf{Parameter Name}} & \multicolumn{2}{c}{\textbf{Experiment Name}}                                    \\ \hhline{~~--}
       &                           & \multicolumn{1}{c|}{\textbf{Area/Density}}      & \textbf{Temporal Granularity} \\ \hline
DC     & (B, A, L, E, lr)          & \multicolumn{1}{c|}{(10, 64, 3, 128, 5e-4)}     & (10, 64, 3, 128, 1e-4)        \\ \hline
MiST                            & (Hsd, E, A, B, lr, Ep)                   & \multicolumn{1}{c|}{(32, 32, 32, 64, 1e-3, 150)} & (32, 32, 32, 32, 1e-3, 200) \\ \hline
CF     & (B, Rl, Ds, Drl, lr, Dr)  & \multicolumn{1}{c|}{(64, 3, 2, 64, 1e-2, 0.1)}  & (64, 6, 4, 64, 1e-2, 0.1)     \\ \hline
HAGEN  & (Rl, Drl, Ss, Sr, lr, Dr) & \multicolumn{1}{c|}{(3, 64, 50, 3, 1e-2, 0.1)}  & \multicolumn{1}{c}{(3, 64, 50, 3, 1e-2, 0.1)}         \\ \hline
ST-SHN & (B, H, lr, Dr)       & \multicolumn{1}{c|}{(16, 128, 1e-3, 0.96)} & (32, 128, 1e-3, 0.96)    \\ \hline
ST-HSL & (B, Cl, Ks, H, lr)        & \multicolumn{1}{c|}{(16, 4, 3, 128, 1e-3)}     & (32, 4, 3, 128, 1e-3)        \\ \hline
AIST   & (B, Rt, Hsl, Ep, lr)      & \multicolumn{1}{c|}{(42, 20, 40, 180, 1e-3)}   & \multicolumn{1}{c}{(42, 20, 40, 120, 1e-3)}         \\ \hline
\end{tabular}
\end{table*}

\subsection{Evaluation Metrics}
We use mean average error (MAE) and rooted mean square error (MSE) to evaluate the prediction models on the regression task. \begin{align*}
\text{MAE} = \frac{1}{n}\sum_{i=1}^{n}|y_i - \hat{y}_i| \qquad
\text{RMSE} = \sqrt{\frac{1}{n}\sum_{i=1}^{n}(y_i - \hat{y}_i)^2} \end{align*}

We use Macro-F1 and Micro-F1 to evaluate the prediction models on the classification task.

\begin{align*}
\text{Micro-F1} = \frac{1}{J}\sum_{j=1}^{J}\frac{2TP_j}{2TP_j+FN_j+FP_j}
\\
\text{Macro-F1} = \frac{2\sum_{j=1}^{J}TP_j}{2\sum_{j=1}^{J}TP_j+\sum_{j=1}^{J}FN_j+\sum_{i=j}^{J}FP_j}
\end{align*}

Here, Here, $n$ represents the number of predictions, $y_i$ represents the ground truth and $\hat{y}_i$ represents the  predicted result., whereas $J=2$ represents the total number of classes and $TP_j, FP_j \, and \, FN_j$ denote the number of true positive, false positive, false negative values in each class, respectively.

\subsection{Evaluation Criteria}
\label{ssec:eval_crit}
Our experiments are categorized into three criteria, which help us assess the performance of the models under under specific conditions and determining their performance. These criteria not only help identify models that are more likely to perform better but also provide insights into the most suitable architectures for different scenarios. The three criteria are as follows.
    
\textit{Evaluation based on area.} Chicago has 77 communities with different sizes from very large (O'Hare - 34.55 sq km) to very small (Oakland - 1.5 sq km). Our aim is to evaluate if geographical properties, i.e, area has an impact on the prediction performance of these deep learning models

\textit{Evaluation based on crime density.}  While it is generally true that larger areas tend to have higher crime rates, there maybe exceptions to this observation as well. In some cases, smaller areas with significant points of interest can exhibit a high density of crimes. Therefore, we introduce a new criterion, crime density, to assess the performance of our models in capturing these variations. To compute the crime density of a community, we divide the total number of crimes by its area. Formally, 
$$\text {Crime  density  of  a  community} =  \frac{\text{\#Crimes of the community}}{\text{Area  of  the  community}}$$

\textit{Evaluation on prediction interval.} The models are evaluated on different granularity of prediction time intervals for all Chicago communities. We try to evaluate how accurate the models' prediction performance is under under different required temporal precision.

\section{Experiment Results}
\label{sec:performance}

\subsection{Evaluation on the Area of the Target Region}
\label{ssec:grp_area}
In this experiment, we aim to assess the impact of the target region's size while predicting crimes. We further want to compare the effectiveness of these models when the size of the target regions vary and identify the models that excel across all different sized target regions. A thorough investigation of the 77 communities in Chicago was carried out to identify areas where the number of communities in each division closely aligns. Five groups were chosen to maintain consistent area ranges while ensuring close community counts within each division. The five groups based on area are as follows: (a) very small, (b) small, (c) medium, (d) large, and (e) very large. Refer to Table~\ref{tab:area_group} for the grouping criteria.

\begin{table}[]
\centering
\setlength{\abovecaptionskip}{-1pt}
\caption{Grouping criteria of the Chicago communities based on area.}
\label{tab:area_group}
\small
    \begin{tabular}{c|c|c|c|}
        \hline
        \textbf{Group} & \textbf{Area Range (Sq. km)} & \textbf{\#Communities} & \textbf{\#Crimes} \\ \hline
        Very Small & $<4$   & 13 & 18070 \\ \hline
        Small      & $4-6$  & 17 & 47813 \\ \hline
        Medium     & $6-8$  & 15 & 52979 \\ \hline
        Large      & $8-10$ & 18 & 78933 \\ \hline
        Very Large & $>10$  & 14 & 63409 \\ \hline
    \end{tabular}
\end{table}                                               

\subsubsection{Performance Comparison for Regression Task}
\begin{itemize}
    \item It is evident from table \ref{tab:area5_reg} that the models tend to perform worse as the community size increases. With the increasing community size, the number of crimes increase too (Refer to Table \ref{tab:area_group}). With the increasing number of crimes, it becomes difficult for the models to predict the exact number of crimes. This can be the reason the models' gradually worse performance for larger areas. 
    
    \item Table~\ref{tab:area5_reg} shows that AIST and HAGEN are the best performing models across all 5 groups. AIST performs better than all the other models including HAGEN with regards to the MAE metric. This suggests that the graph attention layer used to model the spatial dependencies and capture the crime-external feature interaction captures the crime distribution quite well. However, when it comes to capturing the sudden spikes of crime distribution, HAGEN’s Homophily-aware Graph Diffusion Convolution architecture performs comparatively better. It can be attributed to the fact HAGEN not only captures the spatial dependencies of the neighboring regions, it further utilizes the spatial embedding of the regions with same crime profile. On the contrary, AIST only considers the spatial dependencies of the important neighboring regions ignoring the distant regions. 
    
    
    AIST outperforms HAGEN in the MAE metric for all area categories. However, as the area size increases, HAGEN outperforms AIST in the RMSE metric. This suggests that AIST is less sensitive to outliers, with its errors being more uniformly distributed and not having extreme values that significantly skew the error metric. On the other hand, HAGEN's better performance in RMSE indicates that it handles large errors better. This model likely has fewer large outliers or predicts more accurately most of the time but occasionally makes larger mistakes. HAGEN's better performance can be attributed to the fact HAGEN not only captures the spatial dependencies of the neighboring regions, it further utilizes the spatial embedding of the regions with similar profile. On the contrary, AIST only considers the spatial dependencies of the important neighboring regions ignoring the distant regions.
    
    
    \item CrimeForecaster stands out as the next best model across these different categories. CrimeForecaster uses the same graph diffusion convolution network architecture as HAGEN to capture the spatial dependencies, however it does not consider the homophily learning approach thus ignoring the distant regions. This explains the downgrade in performance of CrimeForecaster compared to HAGEN for the RMSE metric. 
    
    \item ST-SHN’s use of hypergraphs to address spatial dependencies does not yield better performance for small sized communities. But as the area gets larger, i.e., the number of crimes increase, it performs comparatively better.  
    
    \item ST-HSL applies a self-supervised learning mechanism to backup its performance when the data is sparse. This makes the model invariant to area or density. 
    
    \item 
    For MiST, the MAE is relatively stable across different area sizes, indicating consistent performance. The RMSE is relatively stable but increases slightly in medium and very large areas.
    
    \item 
    DeepCrime models spatial dependency using an MLP network, which fails to capture the spatial correlations among regions as effectively as the more advanced spatial modules in other models. Consequently, as the area of the target region increases, its performance gradually deteriorates.
\end{itemize}

\begin{center}
    \begin{table}[h]
    \setlength{\abovecaptionskip}{-0.5pt}
    \caption{Regression metrics for groups based on area.} 
        \centering
        \small
        \begin{tabular}{c|c|c|c|c|c|c}
        \hline
        \textbf{Model} & \textbf{Criteria} & \textbf{Very Small} & \textbf{Small} & \textbf{Medium} & \textbf{Large} & \textbf{Very Large} \\ \hline
        \multirow{2}{*}{DC}  & MAE & 0.80 & 1.08 & 1.32 & 1.36 & 1.49 \\ \hhline{~------}
        & RMSE & 1.03 & 2.05 & 2.60 & 2.10 & 2.60 \\ \hline
        \multirow{2}{*}{MiST} & MAE & 0.85 & 0.73 & 0.76 & 0.74 & 0.73 \\ \hhline{~------}
        & RMSE & 0.92 & 0.85 & 0.87 & 0.86 & 0.86 \\ \hline
        \multirow{2}{*}{CF} & MAE & 0.58 & 0.54 & 0.54 & 0.53 & 0.54 \\ \hhline{~------}
        & RMSE & 0.66 & 0.61 & 0.62 & 0.60 & 0.61 \\ \hline
        \multirow{2}{*}{HAGEN} & MAE & 0.46 & 0.50 & 0.54 & 0.52 & 0.51 \\ \hhline{~------}
        & RMSE & 0.52 & \textbf{0.55} & \textbf{0.54} & \textbf{0.55} & \textbf{0.55} \\ \hline
        \multirow{2}{*}{ST-SHN} & MAE & 0.92 & 0.65 & 0.77 & 0.68 & 0.63 \\ \hhline{~------}
        & RMSE & 1.34 & 0.88 & 1.53 & 0.99 & 0.80 \\ \hline
        \multirow{2}{*}{ST-HSL} & MAE & 1.00 & 1.01 & 1.01 & 1.01 & 1.02 \\ \hhline{~------}
        & RMSE & 1.02 & 1.05 & 1.03 & 1.04 & 1.04 \\ \hline
        \multirow{2}{*}{AIST} 
        & MAE & \textbf{0.10} & \textbf{0.35} & \textbf{0.38} & \textbf{0.39} & \textbf{0.42} \\ \hhline{~------}
        & RMSE & \textbf{0.35} & 0.77 & 0.82 & 0.83 & 0.90 \\ \hline
    \end{tabular}
    \label{tab:area5_reg}
    \end{table}
\end{center}



\subsubsection{Performance Comparison for Classification Task}

\begin{itemize}
    \item Contrary to our findings for regression task Table~\ref{tab:area5_cls} shows that models tend to perform better as the community size increases. As evident from Table~\ref{tab:area_group} the number of crimes occurrences are comparatively larger in big communities. In larger areas, crime patterns might be more aggregated, making it easier for models to identify and classify crime hotspots. The larger number of crimes provides more data points, leading to better training and higher performance in classification. On the other hand, regression tasks require predicting exact numbers, which is more sensitive to outliers and noise in the data. In larger areas, the complexity and variability increase, making it difficult for models to capture the exact relationships. 
    \item AIST performs the best across all the groups for both metrics, Macro-F1 and Micro-F1. This is due to the usage of graph attention layers to capture the spatial dependencies and interaction between crime and external features as well as separate temporal modules to capture different trends, i.e., recent, daily and weekly. 
    
    \item CrimeForecaster and HAGEN exhibit competitive performance, with CrimeForecaster generally performing slightly better in predicting crime occurrences indicating the ineffectiveness of homophily-aware learning paradigm introduced in HAGEN. 
    
    \item DeepCrime, originally designed for the classification task surprisingly exhibits better performance with its primitive architectures implying the significance of capturing the temporal dynamics and external feature interactions for the classification task.
\end{itemize}
\vspace{3pt}
\noindent\fbox{%
    \parbox{0.98\columnwidth}{%
        \textit{\textbf{Findings.}} For regression task, AIST is the best performing model across all the groups in terms of the MAE score. In terms of RMSE score, it achieves the best performance for smaller areas with very limited crime occurrences. In contrast, HAGEN, performs the best in terms of RMSE score for larger areas with larger crime occurrences. For classification task, AIST is the winner across all the groups for both Macro-F1 and Micro-F1 metrics.
    }%
}

\begin{table}[ht]
    \centering
    \small
    \setlength{\abovecaptionskip}{-0.5pt}
    \caption{Classification metrics for groups based on area.}
    \begin{tabular}{c|c|c|c|c|c|c}
        \hline
        \textbf{Model} & \textbf{Criteria} & \textbf{Very Small} & \textbf{Small} & \textbf{Medium} & \textbf{Large} & \textbf{Very Large} \\ \hline
        \multirow{2}{*}{DC} & Macro-F1 & 0.24 & 0.30 & 0.33 & 0.38 & 0.42 \\ \hhline{~------}
        & Micro-F1 & 0.36 & 0.41 & 0.55 & 0.56 & 0.59  \\ \hline
        
        \multirow{2}{*}{MiST} & Macro-F1 & 0.18 & 0.22 & 0.28 & 0.31 & 0.35  \\ \hhline{~------}
        & Micro-F1 & 0.21 & 0.34 & 0.39 & 0.42 & 0.45  \\ \hline
        
        \multirow{2}{*}{CF} & Macro-F1 & 0.20 & 0.39 & 0.38 & 0.47 & 0.43  \\ \hhline{~------}
        & Micro-F1 & 0.23 & 0.45 & 0.45 & 0.53 & 0.51  \\ \hline
        
        \multirow{2}{*}{HAGEN} & Macro-F1 & 0.25 & 0.36 & 0.37 & 0.42 & 0.41  \\ \hhline{~------}
        & Micro-F1 & 0.27 & 0.39 & 0.41 & 0.45 & 0.44  \\ \hline
        
        \multirow{2}{*}{ST-HSL} & Macro-F1 & 0.39 & 0.37 & 0.29 & 0.32 & 0.38  \\ \hhline{~------}
        & Micro-F1 & 0.44 & 0.48 & 0.43 & 0.47 & 0.60  \\ \hline
        
        \multirow{2}{*}{AIST} & Macro-F1 & \textbf{0.46} & \textbf{0.50} & \textbf{0.48} & \textbf{0.52} & \textbf{0.61}  \\ \hhline{~------}
        & Micro-F1 & \textbf{0.54} & \textbf{0.60} & \textbf{0.68} & \textbf{0.77} & \textbf{0.73} \\ \hline
    \end{tabular}
    \label{tab:area5_cls}
\end{table}


\subsection{Evaluation on the Crime Density of the Target Region}
\label{ssec:grp_den}
Depending on the socio-economic factors associated with individual communities, even a smaller region can face a large number of crimes, whereas a large but sparsely populated community can hardly face any crimes. By dividing the communities based on crime density, we aim to evaluate how limited as well as ample number of crime records impact the prediction performance of these models. We conducted an analysis of the Chicago communities and divided them into five groups ((a) very low, (b) low, (c) medium, (d) high and (e) large) with equally distributed crime densities, each containing a moderate number of communities. Refer to Table~\ref{tab:density_group} for the categorization criteria.
\begin{table}
    \centering
    \setlength{\abovecaptionskip}{-1pt}
    \caption{Grouping criteria of the Chicago communities based on crime density.}
    \label{tab:density_group}
    \small
    \begin{tabular}{c|c|c}
    \hline
        \textbf{Group} & \textbf{Density Range (crime per sq. km)} & \textbf{\#Communities} \\ \hline
        Very low & $<150$ & 12 \\ \hline
        Low & $150-300$ & 16 \\ \hline
        Medium & $300-450$ & 19 \\ \hline
        High & $450-600$ & 9 \\ \hline
        Very high & $>600$ & 21 \\ \hline
    \end{tabular}
\end{table}
\begin{table}[h]
\caption{Regression metrics for groups based on crime density.}
        \label{tab:density5_reg}
    \centering
    \small
    \begin{tabular}{c|c|c|c|c|c|c}
    \hline
        \textbf{Model} & \textbf{Criteria} & \textbf{Very low} & \textbf{Low} & \textbf{Medium} & \textbf{High} & \textbf{Very high} \\ \hline
    \multirow{2}{*}{DC}  
    & MAE & 0.75 & 0.76 & 0.86 & 1.20 & 1.68 \\ \hhline{~------}
    & RMSE & 0.91 & 0.97 & 1.15 & 1.84 & 2.65 \\ \hline
    \multirow{2}{*}{MiST} 
    & MAE & 0.91 & 0.87 & 0.70 & 0.69 & 0.67 \\ \hhline{~------}
    & RMSE & 0.95 & 0.93 & 0.84 & 0.83 & 0.82 \\ \hline
    \multirow{2}{*}{CF} 
    & MAE & 0.57 & 0.56 & 0.54 & 0.53 & 0.52 \\ \hhline{~------}
    & RMSE & 0.65 & 0.65 & 0.62 & 0.60 & 0.58 \\ \hline
    \multirow{2}{*}{HAGEN} 
    & MAE & 0.46 & 0.50 & 0.50 & \textbf{0.50} & \textbf{0.50} \\ \hhline{~------}
    & RMSE & 0.52 & 0.54 & \textbf{0.55} & \textbf{0.54} & \textbf{0.55} \\ \hline
    \multirow{2}{*}{ST-SHN} 
    & MAE & 1.00 & 0.83 & 0.71 & 0.63 & 0.66  \\ \hhline{~------}
    & RMSE & 1.18 & 1.54 & 1.08 & 0.93 & 0.91 \\ \hline
    \multirow{2}{*}{ST-HSL} 
    & MAE & 1.01 & 1.01 & 1.01 & 1.01 & 1.01 \\ \hhline{~------}
    & RMSE & 1.02 & 1.03 & 1.03 & 1.05 & 1.06 \\ \hline
    \multirow{2}{*}{AIST} 
    & MAE & \textbf{0.11} & \textbf{0.17} & \textbf{0.24} & 0.56 & 0.63 \\ \hhline{~------}
    & RMSE & \textbf{0.36} & \textbf{0.47} & 0.57 & 0.87 & 1.15 \\ \hline
\end{tabular}
\end{table}

\subsubsection{Performance Comparison for Regression Task}
\begin{itemize}
    \item With increasing crime density, models exhibit a somewhat worse performance. As the crime density increases, it becomes difficult for the models to predict the exact number of crimes that is likely to occur. Hence, the models show a greater error when trying to predict for higher crime density communities.
    \item Among the crime prediction models evaluated, AIST and HAGEN are the best performers. Both models use external features and complex architectures that are capable of capturing cross-region and cross-temporal relationships. Particularly, HAGEN exhibits consistent performance across all density groups and shows better performance in high-dense regions. HAGEN utilizes a graph with regions of similar features connected together to effectively capture crime embeddings. So, high density regions can better influence each other that results in improved predictions.
    \item CrimeForecaster ranks as the next best performing model. It captures cross-region dependencies and temporal correlations but does not use external features, resulting in relatively poorer performance compared to the other models.
    \item ST-SHN is the only model that performs better in high density regions compared to low density regions. This model uses a hypergraph-based approach to capture the relationship between one region and all other regions. It also  incorporates cross-category dependencies. As a result, it obtains better embeddings for higher density regions which lead to improved performance.
    \item Although MiST attempts to capture spatial, temporal, and categorical co-dependencies, it relies on simple attention and LSTM architectures. Additionally, MiST does not consider any external features in its predictions, which results in it lagging behind other models in terms of performance. 
    \item ST-HSL exhibits consistency across all density groups. Its dual-stage self-supervised learning algorithm addresses the sparsity issue inherent in crime data. Its hypergraph global dependency model also handles the skewed distribution of crime occurrence at geographical regions. These architectures allow the model to maintain performance regardless of regional density.
    \item DeepCrime, with its basic MLP network for modeling spatial-categorical dependencies, performs the poorest among all the models.

    \item From Figure \ref{fig:hagen_aist_reg}, it is evident that HAGEN's RMSE and MAE values are very consistent across all area and density classes, with little variation. This indicates that HAGEN performs reliably regardless of the area size and crime density. Both RMSE and MAE values for HAGEN are relatively low and constant, suggesting that HAGEN is an accurate model for predicting across different area sizes and densities. AIST shows an increasing trend in both RMSE and MAE as the area size and crime density increases. This indicates that AIST's performance degrades slightly as the area size or crime density becomes larger. AIST's RMSE values are higher than its MAE values, especially noticeable in larger area sizes or densities. This suggests that AIST is more sensitive to outliers, and has some larger prediction errors that significantly affect the RMSE more than the MAE. 
    

\begin{figure}[h]
    \caption{Comparison of regression performance between HAGEN and AIST}
    \label{fig:hagen_aist_reg}
  \begin{subfigure}{.5\textwidth}
  \centering
    \includegraphics[width=1\linewidth]{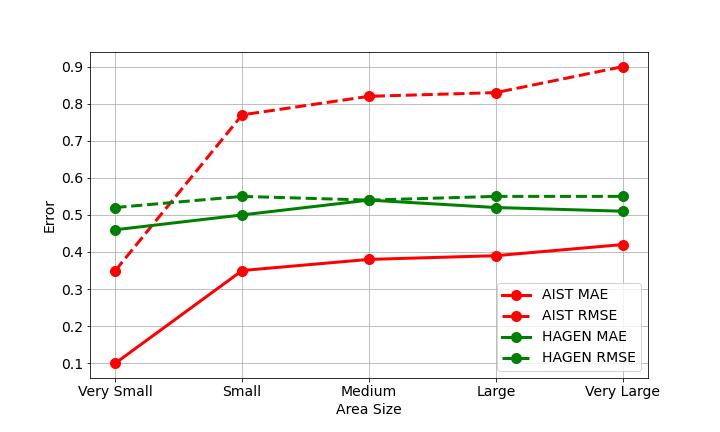}
    \caption{Errors vs Area Size}
  \end{subfigure}%
  \begin{subfigure}{.5\textwidth}
  \centering
    \includegraphics[width=1\linewidth]{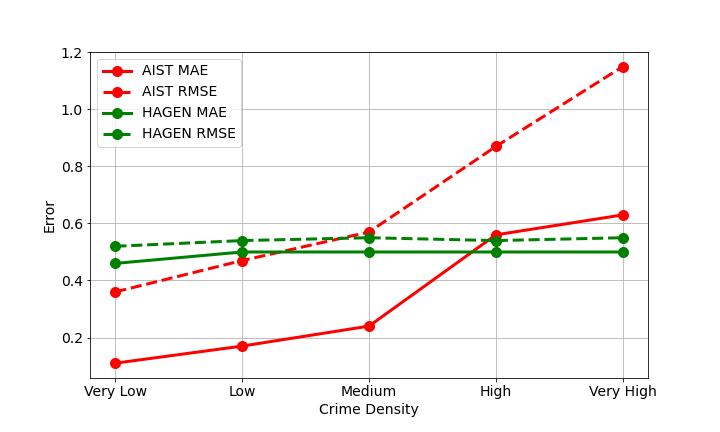}
    \caption{Errors vs Crime Density}
  \end{subfigure}
\end{figure}


\end{itemize}
\begin{table}[!ht]
\small
\centering
\label{tab:density5_class}
\caption{Classification metrics for groups based on crime density.}
\begin{tabular}{c|c|c|c|c|c|c}
    \hline
        \textbf{Model} & \textbf{Criteria} & \textbf{Very low} & \textbf{Low} & \textbf{Medium} & \textbf{High} & \textbf{Very high} \\ \hline
    \multirow{2}{*}{DC} & Micro-F1 & 0.22 & 0.25 & 0.28 & 0.31 & 0.35 \\\hhline{~------}
    & Macro-F1 & 0.26 & 0.29 & 0.38 & 0.44 & 0.49 \\ \hline
    \multirow{2}{*}{MiST} & Micro-F1 & 0.15 & 0.15 & 0.30 & 0.3 & 0.45 \\ \hhline{~------}
    & Macro-F1 & 0.21 & 0.23 & 0.26 & 0.31 & 0.38 \\ \hline
    \multirow{2}{*}{CF} & Micro-F1 & 0.20 & 0.25 & 0.35 & 0.44 & 0.60 \\ \hhline{~------}
    & Macro-F1 & 0.24 & 0.30 & 0.43 & 0.5 & 0.65 \\ \hline
    \multirow{2}{*}{HAGEN} & Micro-F1 & 0.26 & 0.28 & 0.37 & 0.37 & 0.50 \\ \hhline{~------}
    & Macro-F1 & 0.29 & 0.32 & 0.39 & 0.40 & 0.52 \\ \hline
    \multirow{2}{*}{ST-HSL} & Micro-F1 & 0.37 & 0.37 & 0.36 & 0.33 & 0.32 \\ \hhline{~------}
    & Macro-F1 & 0.57 & \textbf{0.58} & \textbf{0.57} & 0.52 & 0.47 \\ \hline
    \multirow{2}{*}{AIST} & Micro-F1 & \textbf{0.45} & \textbf{0.48} & \textbf{0.46 }& \textbf{0.51} & \textbf{0.63} \\ \hhline{~------}
    & Macro-F1 & \textbf{0.65} & 0.57 & 0.56 & \textbf{0.73} & \textbf{0.86} \\ \hline
\end{tabular}
\end{table}

\subsubsection{Performance Comparison for Classification Task}
\begin{itemize}
    \item Contrary to our observation for regression task, for classification the models' performance get better as crime density increases. With higher crime density, there is more data for the models to incorporate the spatio-temporal and external factors to predict the occurrence of a crime. For this reason, although predicting the exact number of crime becomes more difficult, the prediction of the mere occurrence of a crime becomes easier. 
    
    \item AIST emerges as the best performing model for predicting the occurrence of crimes across all scenarios. AIST incorporates multiple external features to effectively capture spatial and temporal dependencies, resulting in accurate predictions. Thus, it performs well for both regression and classification approaches.

    \item In communities characterized by low and medium crime density, ST-HSL performs comparably well to AIST. Its self-supervised learning mechanism for sparse data allows it to provide reliable predictions of crime occurrences when crime data is limited, hence it does comparatively better when crime data is less dense. However, as crime density increases, other models tend to outperform ST-HSL.
    
    \item For high and very high density communities, CrimeForecaster stands out as the second-best model for predicting crime occurrences. With its original classification-based approach and the utilization of DCGRU architecture, CrimeForecaster effectively captures spatial and temporal dependencies, leading to accurate predictions of crime occurrences.
    
    \item HAGEN, an improvement over CrimeForecaster, shows changing performance depending on the density of the region. It performs worse in low-density regions but surpasses CrimeForecaster for high-density regions. This is because of HAGEN's adaptive learning of region dependencies using similar regions as connected nodes in a graph. In high-density regions, HAGEN benefits from more available data, allowing it to learn features more effectively.
    
    \item DeepCrime and MiST perform quite poorly on predicting crime occurrences compared to others. Although these models’ performances tend to get better as the crime density increases, other models do even better than them. DeepCrime models the spatial correlation with an MLP networks, and MiST neglects external features. As crime density gets higher, these factors become more crucial in predicting occurrence of crimes, hence the models perform poorly.

\end{itemize}
\vspace{3pt}
\noindent\fbox{%
    \parbox{0.98\columnwidth}{%
        \textit{\textbf{Findings.}} For regression task, when the crime data is sparse, i.e., for regions with very low to medium crime density, AIST performs best in terms of both MAE and RMSE score. On the contrary, when the number of available crime records is high, i.e., for regions with high to very high density HAGEN performs best. However, for classification task, AIST is the sole best performer.
    }%
}


\subsection{Evaluation on the Temporal Granularity of Prediction}
\label{ssec:grp_time}
Crime forecasting models can be trained to predict crimes at different temporal granularity, i.e., hourly, daily, weekly and so on. We evaluate the models based on the following four temporal precision: (a) 4 hours, (b) 6 hours, (c) 12 hours, and (d) 24 hours. Our aim is to evaluate how accurate the models are in predicting crimes at different temporal precision and whether the temporal precision adversely affects a model's prediction performance or not. 


\begin{table}[!ht]
\centering
\small
\caption{Regression metrics for different temporal granularity.}
\label{tab:tem_reg}
\begin{tabular}{c|c|c|c|c|c}
\hline
    \textbf{Model} & \textbf{Criteria} & \textbf{4h} & \textbf{6h} & \textbf{12h} & \textbf{24h} \\ \hline
    \multirow{2}{*}{DC} & MAE & 0.88 & 0.93 & 0.99 & 1.22 \\ \hhline{~-----}
    & RMSE & 0.87 & 0.91 & 0.89 & 1.68 \\ \hline
    \multirow{2}{*}{MiST} & MAE & 0.37 & 0.49 & 0.57 & 0.76 \\ \hhline{~-----}
    & RMSE & 0.54 & 0.57 & 0.67 & 0.87 \\ \hline
    \multirow{2}{*}{CF} & MAE & 0.59 & 0.53 & 0.55 & 0.54 \\ \hhline{~-----}
    & RMSE & 0.71 & 0.65 & 0.64 & 0.62 \\ \hline
    \multirow{2}{*}{HAGEN} & MAE & 0.50 & 0.49 & 0.50 & 0.49 \\ \hhline{~-----}
    & RMSE & \textbf{0.53} & \textbf{0.53} & \textbf{0.54} & \textbf{0.54} \\ \hline
    \multirow{2}{*}{ST-HSL} & MAE & 1.02 & 1.00 & 1.02 & 1.01 \\ \hhline{~-----}
    & RMSE & 1.03 & 1.02 & 1.03 & 1.04 \\ \hline
    \multirow{2}{*}{AIST} & MAE & \textbf{0.34} & \textbf{0.35} & \textbf{0.41} & \textbf{0.44} \\ \hhline{~-----}
    & RMSE & 0.57 & 0.61 & 0.65 & 0.67 \\ \hline
\end{tabular}
\end{table}


\subsubsection{Performance Comparison for Regression Task}
\begin{itemize}
    \item A general trend here is that as the temporal granularity becomes coarser the models' performance deteriorate (Refer to Table \ref{tab:tem_reg}). 
    This is down to the fact that at finer granularity, the number of crimes in adjacent time steps remain more consistent compared to when the temporal granularity is coarser. Thus, it becomes hard for the models to predict crime accurately.
    
    \item Similar to previous observations, AIST performs the best across all temporal granularity for the MAE metric. In all time intervals AIST does well in predicting number of crimes. But it comes second best for the RMSE metric. The best performing model across all the temporal granularity for RMSE metric is HAGEN. This is due to the homophily-aware architecture that HAGEN provides which makes it easier for HAGEN to capture sudden spikes.
    
    \item CrimeForecaster, HAGEN, ST-HSL show consistent performance across different temporal granularity. Despite having to predict larger crime numbers with coarser temporal granularity, these models benefit from the fact that the actual number of zero crimes is small. DeepCrime performs well at finer resolution compared to when the temporal resolution is coarser.

\end{itemize}

\subsubsection{Performance Comparison for Classification Task}
\begin{itemize}
    \item As the temporal resolution becomes coarser, the prediction performance of all the models improve which suggests that for classification task it is harder for models to predict at a finer temporal resolution (Refer to Table \ref{tab:time_macro_micro}). This can be attributed to the fact that crimes exhibit strong daily, weekly, monthly and seasonal correlation. With coarser temporal resolution these trends become more obvious for the models to capture. 
    \item AIST outperforms all the other models in terms of both the Macro-F1 and Micro-F1 score at different temporal granularity. The performance gap of AIST is more prominent while predicting at finer temporal granularity. This is due to the fact that contrary to the other models AIST explicitly captures the recent, daily and weekly trends in crime data by using three different LSTM modules. 
    \item In contrast to its performance in other experiments, HAGEN performs worse, more so at finer temporal granularity. The adaptive graph architecture utilized by HAGEN struggles to effectively learn and adapt when the available feature information becomes limited.
    
    \item Contrary to the regression task, for classification DeepCrime shows a strong performance when the temporal granularity of the prediction is coarser. Classification, being a comparatively simpler task than regression, allows DeepCrime to leverage its simple design and effectively predict the crime occurrences at coarser temporal precision.
    
\end{itemize}
\vspace{3pt}
\noindent\fbox{%
    \parbox{0.98\columnwidth}{%
        \textit{\textbf{Findings.}} For regression task, it is harder for the models to predict crimes at coarser temporal granularity. AIST performs best in terms of the MAE score, while HAGEN performs best in terms of the RMSE score. Contrary to regression, for classification task it is harder for the models to predict crimes at finer temporal granularity. AIST is the best performing model for both Macro-F1 and Micro-F1 metrics for classification.
    }%
}

\begin{center}
    \begin{table}[!ht]
    \centering
    \small
    \caption{Classification metrics for different temporal granularity.}
    \label{tab:time_macro_micro}
    \begin{tabular}{c|c|c|c|c|c}
    \hline
        \textbf{Model} & \textbf{Criteria} & \textbf{4h} & \textbf{6h} & \textbf{12h} & \textbf{24h} \\ \hline
        \multirow{2}{*}{DC} & Macro-F1 & 0.24 & 0.32 & 0.34 & 0.42 \\ \hhline{~-----}
        & Micro-F1 & 0.24 & 0.38 & 0.40 & 0.55 \\ \hline
        \multirow{2}{*}{MiST} & Macro-F1 & 0.29 & 0.35 & 0.38 & 0.32 \\ \hhline{~-----}
        & Micro-F1 & 0.30 & 0.38 & 0.41 & 0.37 \\ \hline
        \multirow{2}{*}{CF} & Macro-F1 & 0.16 & 0.21 & 0.30 & 0.38 \\ \hhline{~-----}
        & Micro-F1 & 0.20 & 0.25 & 0.35 & 0.44 \\ \hline
        \multirow{2}{*}{HAGEN} & Macro-F1 & 0.09 & 0.13 & 0.23 & 0.37 \\ \hhline{~-----}
        & Micro-F1 & 0.10 & 0.15 & 0.25 & 0.40 \\ \hline
        \multirow{2}{*}{ST-HSL} & Macro-F1 & 0.13 & 0.23 & 0.34 & 0.34 \\ \hhline{~-----}
        & Micro-F1 & 0.16 & 0.32 & 0.51 & 0.54 \\ \hline
        \multirow{2}{*}{AIST} & Macro-F1 & \textbf{0.45} & \textbf{0.48} & \textbf{0.50} & \textbf{0.51} \\ \hhline{~-----}
        & Micro-F1 & \textbf{0.78} & \textbf{0.59} & \textbf{0.64} & \textbf{0.68} \\ \hline
    \end{tabular}
\end{table}
\end{center}

\subsection{Exploring the Impact of Area Size with Fixed Crime Density}
The aim of these experiments is to explore the impact of different community areas on performance when crime density remains constant. To accomplish this, we select a specific density group and divide the communities in that group into five categories based on their area size: very small, small, medium, large, and very large. Next, we calculate the mean of the metrics for communities for each of the five area categories. We then compute the variance of the five means for that density group, which gives us the variance of performance among the different area groups, while holding density constant. Our findings indicate that the majority of the models demonstrate minimal variances, typically ranging from 1e-2 to 1e-6. This indicates that for a given density category, there is minimal variation in the performance of communities with different sizes.

\begin{table}[]
\centering
\caption{Impact of various spatial components on performance}
\label{tab:imp_spa_tab}
\begin{tabular}{c|ccc|cc|cc}
\hline
\multirow{2}{*}{Model} &
  \multicolumn{3}{c|}{Spatial Correlation On} &
  \multicolumn{2}{c|}{Regression Metric} &
  \multicolumn{2}{c}{Classification Metric} \\ \cline{2-8} 
 &
  \multicolumn{1}{c|}{Crime} &
  \multicolumn{1}{c|}{External Data} &
  Extent &
  \multicolumn{1}{c|}{MAE} &
  RMSE &
  \multicolumn{1}{c|}{Micro F1} &
  Macro F1 \\ \hline
\multirow{4}{*}{HAGEN} &
  \multicolumn{1}{c|}{Y} &
  \multicolumn{1}{c|}{} &
  Y &
  \multicolumn{1}{c|}{0.287} &
  0.226 &
  \multicolumn{1}{c|}{0.707} &
  0.663 \\ \cline{2-8} 
 & \multicolumn{1}{c|}{Y} & \multicolumn{1}{c|}{}  & N & \multicolumn{1}{c|}{0.283} & 0.226 & \multicolumn{1}{c|}{0.706} & 0.663 \\ \cline{2-8} 
 & \multicolumn{1}{c|}{N} & \multicolumn{1}{c|}{}  & Y & \multicolumn{1}{c|}{0.486} & 0.280 & \multicolumn{1}{c|}{0.526} & 0.472 \\ \cline{2-8} 
 & \multicolumn{1}{c|}{N} & \multicolumn{1}{c|}{}  & N & \multicolumn{1}{c|}{0.491} & 0.277 & \multicolumn{1}{c|}{0.492} & 0.473 \\ \hline
\multirow{4}{*}{AIST} &
  \multicolumn{1}{c|}{Y} &
  \multicolumn{1}{c|}{Y} &
   &
  \multicolumn{1}{c|}{0.440} &
  0.670 &
  \multicolumn{1}{c|}{0.680} &
  0.510 \\ \cline{2-8} 
 & \multicolumn{1}{c|}{Y} & \multicolumn{1}{c|}{N} &   & \multicolumn{1}{c|}{1.330} & 2.058 & \multicolumn{1}{c|}{0.674} & 0.496 \\ \cline{2-8} 
 & \multicolumn{1}{c|}{N} & \multicolumn{1}{c|}{Y} &   & \multicolumn{1}{c|}{1.332} & 2.052 & \multicolumn{1}{c|}{0.684} & 0.503 \\ \cline{2-8} 
 & \multicolumn{1}{c|}{N} & \multicolumn{1}{c|}{N} &   & \multicolumn{1}{c|}{1.332} & 3.011 & \multicolumn{1}{c|}{0.674} & 0.496 \\ \hline
\end{tabular}
\end{table}

\subsection{Exploring the Impact of Spatial Components on Crime Prediction Performance}
Previous experiments indicate that HAGEN and AIST consistently outperform other models across various scenarios. In our subsequent experimentation, we focus only on these two models, deactivating different spatial components to explore their relative importance in these models' success.

HAGEN and AIST utilize complex architectures designed to capture various spatial correlations in crime data. We can identify three types of spatial correlations captured in these models: spatial correlation on crime, spatial correlation on external data, and spatial correlation on extent.

HAGEN's Graph Learning Layer leverages POI data to construct a regional graph, capturing spatial correlations of crimes across different regions. The Homophily constraint ensures that neighboring nodes in the regional graph display similar crime patterns, thereby capturing spatial correlation on extent.

AIST incorporates two variants of the Graph Attention Network: hGAT and fGAT. The hGAT component learns region crime embeddings by integrating hierarchical information, capturing the spatial correlation on crime. Meanwhile, the fGAT component embeds external features into the model, capturing the spatial correlation on external data.

We deactivate the components responsible for capturing these three types of spatial correlation to determine their importance in the superior performance of the models. Table \ref{tab:imp_spa_tab} presents the results of our study.

\vspace{3pt}
\noindent\fbox{%
    \parbox{0.98\columnwidth}{%
        \textit{\textbf{Findings.}} When its components for capturing spatial correlation are active, HAGEN performs better in all scenarios. In contrast, when these components are active, AIST performs noticeably better on the regression task and marginally better on the classification task. The findings show that techniques for capturing spatial correlations in crime data improve the models' ability to predict crimes.
    }%
}


\subsection{Exploring the Impact of External Features on Crime Prediction Performance}
\label{ssec:ablation}
To find out the effect of external data on the performance of the models we conduct ablation study on the models that use external datasets: DeepCrime, HAGEN, AIST. Table \ref{tab:ablation} shows the performance of the models with and without using external feature data. DeepCrime takes POI and 311 public service complaints data as external features along with crime data. For both regression and classification metrics, DeepCrime performs slightly better when external data is used. HAGEN uses POI data to build the region graph. Without the POI external feature, HAGEN performs better for regression and negligibly worse for classification. AIST uses POI data and taxi flow data for predicting crimes. With these external features turned off AIST perform worse for both the prediction types. However, for classification the degradation is small.

\vspace{3pt}
\noindent\fbox{%
    \parbox{0.98\columnwidth}{%
        \textit{\textbf{Findings.}} Compared to other models, the introduction of external features have more visible impact on the prediction performance of AIST due to its attempt to explicitly capture the interaction between crime and external data. In contrast, the gain in performance is marginal for models (e.g., DeepCrime, HAGEN) which treat external datasets only as an additional feature ignoring its interaction with crime data.
    }%
}
\begin{table}[htbp]
\centering
\caption{Performance of models in ablation study. (R: Regression, C: Classification) }
\label{tab:ablation}
\small
\begin{tabular}{c|c|c|c|c}
\hline
\textbf{Model}             & \textbf{Task}                   & \textbf{Metric} & \textbf{w External} & \textbf{w/o External} \\ \hline
\multirow{4}{*}{DC} & \multirow{2}{*}{R}     & MAE             & \textbf{0.912}         & 0.916                     \\ \hhline{~~---} 
                           &                                 & RMSE            & \textbf{0.937}         & 0.939                     \\ \hhline{~----} 
                           & \multirow{2}{*}{C} & Macro-F1        & \textbf{0.426}         & 0.420                     \\ \hhline{~~---} 
                           &                                 & Micro-F1        & \textbf{0.549}         & 0.546                     \\ \hline
\multirow{4}{*}{HAGEN}     & \multirow{2}{*}{R}     & MAE             & \textbf{0.493}         & 0.498                     \\ \hhline{~~---} 
                           &                                 & RMSE            & \textbf{0.541}         & 0.547                     \\ \hhline{~----} 
                           & \multirow{2}{*}{C} & Macro-F1        &     \textbf{0.369}           &  0.367          \\ \hhline{~~---} 
                           &                                 & Micro-F1        &\textbf{0.401}                   & 0.396            \\ \hline
\multirow{4}{*}{AIST}      & \multirow{2}{*}{R}     & MAE             & \textbf{0.323}         & 0.331                     \\ \hhline{~~---} 
                           &                                 & RMSE            & \textbf{0.634}         & 0.653                     \\ \hhline{~----} 
                           & \multirow{2}{*}{C} & Macro-F1        & \textbf{0.453}         & 0.451                     \\ \hhline{~~---} 
                           &                                 & Micro-F1        & \textbf{0.777}         & 0.775                     \\ \hline
\end{tabular}
\end{table}



\section{Key Findings} 
\label{sec:findings}

\noindent\textbf{Q1. Which model/models exhibit superior performance in crime regression task?} Table~\ref{tab:tem_reg} (See 24h column) reveals that AIST and HAGEN demonstrate the best performance among the competing models. AIST excels in terms of the MAE metric, indicating its superior performance in predicting the average number of crimes. On the other hand, HAGEN performs best in terms of the RMSE metric, showcasing its ability to capture sudden spikes in crime occurrences. These two models perform at a similar level for overall crime prediction. Their utilization of complex architectures and consideration of multiple external features contribute to their superior performance.

\textbf{Q2. Which model/models exhibit superior performance in crime classification task?} AIST demonstrates superior performance compared to all other models in predicting the occurrence of crimes, as evident from Table \ref{tab:time_macro_micro} (See 24h column). This aligns with its overall better performance in regression tasks. AIST's proficiency in predicting the number of crimes naturally translates to better performance in predicting their occurrence.

\textbf{Q3. Which model/models exhibit superior performance when the temporal granularity of the prediction is very fine, i.e., 4h?} 
AIST, the best performing model in terms of MAE score outperforms its closest competitor MiST by $8.1\%$, whereas HAGEN the best performing model for RMSE score outperforms its closest competitor MiST by $1.85\%$. The performance gap in predicting crimes at a finer granularity is more evident for crime classification task. AIST comprehensively outperforms the next best performing model with an improvement of $55.17\%$ and $160\%$ in terms of Macro-F1 and Micro-F1 metrics, respectively.

\textbf{Q4. Which model/models exhibit superior performance when the crime dataset is very sparse?} 
AIST comes across as the best performing model across both the regression and the classification task for all the evaluation metrics due to the model's design architecture which separately captures different temporal trends as well as the interaction between the external features and the crime data. 

\textbf{Q5. Is the crime data itself enough for crime prediction task in absence of external features?} Table~\ref{tab:ablation} suggests that crime data itself is enough for a good crime prediction model be it for regression or classification task. Even though incorporating external features improve the prediction performance for the respective models, the improvement is marginal. In our experiments we find that incorporating external features improve the prediction performance of DeepCrime, HAGEN and AIST $0.43\%, 1\%, 2.41\%$ in terms of MAE, $0.21\%, 1.09\%, 0.29\%$ in terms of RMSE, $1.43\%, 0.54\%, 0.44\%$ in terms of Macro-F1 and $0.54\%, 1.26\%, 0.26\%$ in terms of Micro-F1, respectively.

\textbf{Q6. Does introducing external features always improve the prediction performance of the crime prediction models?} Analyzing the models such as DeepCrime, HAGEN and AIST that use external features for predicting crimes, we found that introducing external features always improve the prediction performance of the models, be it marginal. Out of these three models only AIST attempts to capture the interaction between the crime data and external features by introducing a graph attention layer. The other two models DeepCrime and HAGEN incorporate anomaly and POI data with the temporal and spatial view by directly concatenating with the temporal and spatial view, respectively ignoring the interaction between the crime and external features. As a result, the improvement in performance of AIST with external features is comparatively better than the other two, i.e., incorporating external features improve AIST's performance $2.41\%$ compared to $0.41\%, 1\%$ for DeepCrime and HAGEN for MAE, respectively.

\textbf{Q7. Is capturing the spatial correlation among different regions actually necessary for crime prediction task?} Out of all the competing models, only DeepCrime ignores the spatial correlation between regions and performs the worst for regression task (Refer to Table~\ref{tab:tem_reg}), suggesting that incorporating spatial correlation of the regions is essential for the regression tasks. However, its superior performance over all the competing models (excluding AIST) for classification task suggests that considering spatial correlation is not a must for the classification task as long as the temporal correlation and external features are integrated into a model's design architecture. Refer to Table~\ref{tab:time_macro_micro}.

\textbf{Q8. Do regression and classification task require designing model architecture with different properties?} Models that leverage complex neural network architecture to capture the spatial dependencies tend to perform better for regression task specially when there is ample crime data, e.g.,  On the other hand, models that explicitly captures different temporal trends tend to perform better for the classification task, e.g., AIST.

\section{Recommendations}
\label{sec:reco}
\textit{Capturing spatio-temporal correlation (R1).} Design crime prediction models such that it can explicitly capture both the spatial correlation among the regions and the temporal correlation of the crime data. For spatial correlation, it is essential to not only capture the dependencies of the neighboring regions, but also the similar regions with same crime profile that may be far away from the target region. As for temporal correlation, it helps to explicitly capture daily, weekly temporal dynamics of the crime data.

\textit{In absence of external features (R2).} In case of the unavailability of the external features, building a model that can utilize the spatial and temporal is good enough for a crime prediction model. A combination of GNN (i.e., GCN, GAT) and RNN (i.e., LSTM, GRUs) for capturing the spatial and temporal correlation is a good starting point while designing the crime prediction models. Avoid using LSTMs for capturing the spatial correlation or convolution networks for capturing the temporal correlation. 

\textit{Utilizing external features (R3).} If external features are available, it is always beneficial to utilize it. However, incorporate explicit modules in your crime prediction model such that it can capture the interaction between the crime data and the external features to  benefit fully from utilizing external features.

\textit{Regression-only task (R4).} While designing crime prediction models only for regression purpose, the spatial module requires special attention since it is the driving force that will decide the prediction performance. The spatial module should be able to capture not only the spatial dependencies of the neighboring regions but also it should be able to learn from the non-neighboring regions that exhibit similar crime patterns to the target region. 

\textit{Classification-only task (R5).} While designing crime prediction models only for classification purpose, utmost importance should be given to capture the different trends of crime. Models should include separate modules to capture recent, daily, weekly, monthly crime trends. Only after the temporal dynamics is fully captured, spatial and external feature modules can be introduced that complement the temporal module.


\section{Conclusion}
\label{sec:conclusion}
In this paper, we have conducted a detailed experimental study of all major modern deep learning based crime prediction models and compared them in a unified environment. As a number of different deep learning based models have been proposed for crime prediction in recent years, and there has been a lack of direct comparisons among these models, researchers and practitioners face hurdles in analyzing and adapting these models. 
Therefore, we have systematically compared the models' performance across different scenarios, including variations in community area size, crime density, and prediction intervals. This experimental study has enabled us to identify the models that performed best in specific scenarios and gain insights into the suitability of different architectures under various conditions. 


\bibliographystyle{unsrtnat}
\bibliography{references}  






\end{document}